# Social preferences with unstable interactive reasoning: Large language models in economic trust games

Jiamin Ou [1*], Emile Eikmans[2], Vincent Buskens[1], Paulina Pankowska[1], and Yuli Shan[3]

[1]Department of Sociology, Utrecht University, Padualaan 14, 3584 CC Utrecht, Netherlands

[2] School of Economics (U.S.E.), Utrecht University, Kriekenpitplein 21-22, 3584 EC Utrecht, Netherlands

[3]School of Geography, Earth, and Environmental Sciences, University of Birmingham, United Kingdom

* j.ou@uu.nl

## Abstract

While large language models (LLMs) have demonstrated remarkable capabilities in understanding human languages, this study explores how they translate this understanding into social exchange contexts that capture certain essences of real-world human interactions. Three LLMs - ChatGPT-4, Claude, and Bard - were placed in economic trust games where players balance self-interest with trust and reciprocity, making decisions that reveal their social preferences and interactive reasoning abilities. Our study shows that LLMs deviate from pure self-interest and exhibit trust and reciprocity even without being prompted to adopt a specific persona. In the simplest one-shot interaction, LLMs emulated how human players place trust at the beginning of such a game. Larger human-machine divergences emerged in scenarios involving trust repayment or multi-round interactions, where decisions were influenced by both social preferences and interactive reasoning. LLMs' responses varied significantly when prompted to adopt personas like selfish or unselfish players, with the impact outweighing differences between models or game types. Response of ChatGPT-4, in an unselfish or neutral persona, resembled the highest trust and reciprocity, surpassing humans, Claude, and Bard. Claude and Bard displayed trust and reciprocity levels that sometimes exceeded and sometimes fell below human choices. When given selfish personas, all LLMs showed lower trust and reciprocity than humans. Interactive reasoning to the actions of counterparts or changing game mechanics appeared to be random rather than stable, reproducible characteristics in the response of LLMs, though some improvements were observed when ChatGPT-4 responded in selfish or unselfish personas.

# 1 Introduction

The advent of ChatGPT and other similar large language models (LLMs) such as ChatGPT, Claude, Bard, and Llama distinguish themselves from their predecessors through their sophisticated understanding of context, nuance, and the subtleties of human language. Based upon a "black-box" like neural network, ChatGPT and other LLMs continue to evolve through training on vast corpora of human-generated data. As human-generated data inherently includes rationale and social traits such as values, norms, and biases that characterize human behavior, a pivotal question emerges: To what extent do LLMs, shaped by extensive human input, emulate not only the linguistic and cognitive capabilities of humans but also human-like social interactions [1,2]?

The primary goal of LLMs – providing human-like responses in conversation – has been largely fulfilled as responses of LLMs are not only grammatically correct, but also often logically sound, knowledgeable, and communicated in a human-like tune that seems to understand and respond to sentiments, emotions, and social cues. They also adhere to certain social norms in communication such as politeness, turn-taking, and appropriate topic selection [3]. The ways LLMs engage in conversation reflect some extent of social intelligence, which is the key to making their responses in conversations human-like. Even though this seemingly social intelligence is based on learned patterns rather than actual reasoning abilities and social awareness, it is intriguing and crucial to understand how far LLMs can emulate human responses beyond conversations, such as decisions made in social exchange contexts. For example, LLMs understand well the meaning of words like fairness, altruism, and reciprocity and are able to explain them in conversations, but in a behavioral context where a decision is made, such as how much to share with others to show fairness, it remains unclear how LLMs could translate the meaning of fairness they learned from the training dataset and apply it to an actual decision. It is also unclear to what extent their responses align with choices made by humans. The answer to this question has substantial implications, such as using these LLMs as substitutes or surrogates of human subjects in questionnaires, behavioral tests, observational studies, and agent-based modeling [1], or its potential as assistants to humans in making profound decisions embedded in specific contexts such as national defense and security.

Compared to the large body of literature on the cognitive and reasoning capacities of LLMs [4–7], studies on how LLMs provide human-like behavioral responses are limited. While cognitive tests generally consist of questions with "correct" answers, behavioral experiments often use social dilemmas without right or wrong actions. A few studies used military simulations and war games to study the responses made by LLMs in interactive non-cooperative scenarios [8–10]. In fictional geopolitical conflicts, high-level agreement was found between LLMs and human responses with significant differences in individual actions and preferences, such as more aggressive actions taken by LLMs [9, 10]. Economic and social psychology experiments were used to study the response of LLMs. The Milgram Shock Experiment, for example, studies obedience to authority by asking participants to shock objects. Higher-level obedience was found in LLMs than in human participants [11]. Regarding fairness and rationality, Aher et al. (2023) studied the response of LLMs in the

Ultimatum Game [11]. Out of the eight earlier models of ChatGPT, one of them showed alignment with human perceptions, seeing 50-100% and 0-10% of the total endowments as fair and unfair offers, respectively. Akata *et al*. (2025) [2] played repeated games such as Prisoner's Dilemma and Battle of the Sexes with LLMs and study their cooperation and coordination behaviors. They found that LLMs perform well at self-interested games but behave suboptimally in games that require coordination.

Among these experimental tools, the economic trust game (hereinafter referred to as 'trust game') is a key tool to capture social preference regarding the expression and repayment of trust, an important social signaling mechanism that influences human behavior and human systems of morality [12]. It encodes trust into economic exchange games in which two players send money back and forth with the risk of losing it. Johnson & Obradovich (2024) [13] used the trust game and showed LLMs responded to actual incentives in trusting behavior. Lore & Heydari (2024) revealed the differences across LLMs in balancing game structure and contextual framing of trust games, with ChatGPT-4 focusing more on internal game mechanics while Llama was more influenced by contextual elements [14]. Xie *et al.* (2024) [15] studied the trust behavior of LLMs by assigning 53 personas with detailed demographics such as name and gender to each LLM and found that LLM agents generally exhibit trust behavior as referred to as *agent trust*. This existing study highlights the potential of trust game in uncovering trust patterns and dynamics in LLMs. However, several aspects remain underexplored, including the neutral position taken by each LLM when no persona is assigned, their perception of general personas without interference from detailed demographic attributes and the adaptive patterns of how agents develop trust over repeated interactions.

Despite previous efforts, there remains a critical gap in understanding behavioral differences between human responses and those generated by LLMs, as well as among different LLM systems themselves in social context. Using the trust game as an example, valuable insights can be offered to compare the amounts sent and returned by LLMs and humans in comparable contexts and examine whether LLMs resemble tendencies of social preferences that align with certain human population segments. In addition, choices made by players in the trust game were driven by social preferences entangled with interactive reasoning on how they foresee the likely actions taken by the other players, learn from past interactions, and adapt to changes in game mechanics. A systematic analysis of both social preference and interactive reasoning between LLMs and humans could provide a more comprehensive understanding of how well LLMs emulate human social interactions and the specific ways in which they do so. To this end, we simulated an experimental design comparable to a previous human laboratory experiment [16] featuring three variants of the trust game and applied it to three mainstream LLMs and their personas. The three trust games were the one-shot game, the repeated game, and the probabilistic repeated game. Specifically, the one-shot game is a single-round trust game that captures immediate trust and reciprocation. The game involves a sender and a receiver, both starting with 12 points. The sender kicks off the game by choosing to send 0, 4, 8, or 12 points to the receiver, which are then tripled. After receiving the additional points from the sender, the receiver can return any amount of their points to the sender. In the repeated game, players engage in six rounds, knowing they will interact with the same partner and     receive feedback on prior actions. The iterative rounds enable players to learn from past interactions and provide room for long-term trust development and cooperation. In the probabilistic repeated game, players participate in up to six rounds of the trust game. However, a six-sided die is rolled after each round, and if a "6" appears, the game ends prematurely. As a close variant of the repeated game, it cross-validates the consistency of LLMs' responses in multi-round interactions.

In the games, each LLM was assigned one of the following three personas: unspecified, selfish, or unselfish. For the unspecified persona, the prompt did not mention any preferences the LLM should consider when playing the game. This reflects the 'natural' or default position taken by an LLM when no specific persona prompt is provided. The selfish persona was explicitly instructed to ask an LLM to play as a selfish player, while an LLM was asked explicitly to act as an unselfish player under the unselfish persona. The contrasting selfish and unselfish personas allow us to explore LLMs' responses when they were asked to operationalize selfishness and unselfishness in specific contexts.

We employed three LLMs–ChatGPT-4 (developed by OpenAI), Claude (created by Anthropic), and Bard (now known as Gemini, developed by Google)–in the experiment. Reactions of LLMs in the games were compared across models and with human participants.

## 2 Methods

This experiment involves three LLMs, namely ChatGPT-4, Claude, and Bard, in three types of economic trust games with unspecified, selfish, and unselfish personas. Responses of three LLMs were compared to those of human participants in similar setups from a previous study [16].

### 2.1 Economic trust games

Three variants of economic trust games– the one-shot game, the repeated game, and the probabilistic repeated game– are used to offer distinct insights into trust and cooperation in economic interactions.

The one-shot game is a single-round trust game that captures immediate trust and reciprocation. The sender starts with 12 points and can send 0, 4, 8, or 12 points to the receiver, which are then tripled. The receiver also starts with 12 points and receives the additional points from the sender, totaling up to 48 points. The receiver can return any amount of their points to the sender. For instance, if the sender transfers 4 points, the receiver gains 12 points. With the initial 12 points, the receiver now has a total of 24 points and can choose to return any amount between 0 and 24 points to the sender.

In the repeated game, players engage in six rounds of the same trust game, knowing they will interact with the same partner and receive feedback on prior actions. In the probabilistic repeated game, players participate in up to six rounds of the trust game. However, after each round, a six-sided die is rolled, and if a six appears, the game ends prematurely. After six rounds, the game ends for sure.

We collected approximately 600 to 700 game responses for the one-shot game for each LLM and persona. Each game consisted of a single-round interaction. For the multi-round games, we analyzed

330 to 500 unique games for each game type, LLM, and persona. Each unique game included up to six rounds of interactions. The exact number of unique games is detailed in Appendix A. To capture the variability in LLMs and ensure the game remains within the memory limit of the LLMs, the outcomes were not collected in a single inquiry. In the one-shot game, each inquiry involved the LLMs playing against themselves 30 times (see Appendix B1.3-5). Regarding the repeated and probabilistic repeated games, each inquiry had the LLMs playing against themselves 5 times, resulting in the collection of game outcomes from 130 separate inquiries (see Appendix B2.3-5&3.3-5).

## 2.2 Details of LLMs and personas

We manually queried ChatGPT-4, Claude, and Bard through their respective webpage chatbots from November 2023 to January 2024. For ChatGPT-4, developed by OpenAI, interactions were conducted via the webpage at chat.openai.com. For Claude, developed by Anthropic, the experiments were performed through the webpage at claude.ai. Another LLM used was Bard, developed by Google, which has since been renamed Gemini; these interactions took place at bard.google.com (now gemini.google.com).

We conducted the experiment in the web versions of LLMs instead of their APIs to observe how LLMs behave when interacting with everyday uses within the fine-tuned environments prescribed by each company. We recognize the advantages of using APIs for experimentation since it allows the users to get into the model and pre-script model parameters such as temperature, to build a well-controlled environment. But by pre-scripting model parameters, LLMs might behave differently than what they would have when they interact with ordinary users in the real world. Given our interest in the trust and cooperation of LLMs emerged in real-world interactions and its comparison to humans, we decided to experiment on the web versions, which might reveal implicit built-in safeguards, biases, ethical considerations and other company-specific features which would not necessarily emerge in a pre-planned API experiment. One of the limitations of this approach, however, is that we cannot provide the exact parameters of each model involved in the experiment. OpenAI, Anthropic and Google did not disclose the model parameters of their web-version LLM, although the values of some parameters such as temperature were believed to be set between 0.7 to 0.8 as defaults. It implies that here we studied the "behavior" of each LLM in a black-box-like manner, observing its decision-making process as if it interacts with ordinary users. To capture the variations of webpage responses, a few hundred repetitions (See 2.1) were performed for each game configuration.

At the beginning of each game, we first introduced the rules of the games to each LLM. After the introduction of rules, we used Chain-of-Thought (CoT) prompting [17] to test if each LLM understands the rules. If a wrong answer was provided, the LLM was shown, with examples, how the game functioned (all the prompts can be found in Appendix B). Given the choice of web version, each LLM naturally retains the memories of previous conversations about game mechanics and corrections.

Following the introduction of game mechanics, LLMs started to play the games with or without personas prompting. First, we asked each LLM to play against themselves without given persona.

This resulted in the data for the unspecified persona. Following that, we provided specific prompts to LLMs, asking them to play the game again while adopting certain personas, such as "individuals with selfish or unselfish behaviors" (see Appendix A1.3-5, 2.3-5&3.3-5). This approach was used to study the game outcomes for selfish and unselfish personas, respectively.

## 2.3 Human data for comparison

We used human data from a previous experiment that utilized similar game settings [16] for comparison with LLMs. The experiment was designed for different purposes, namely to test the effect of testosterone administration, but the setup of six one-shot games versus a six-times repeated game was identical to the ones used for LLMs. It should be noted that this previous experiment used exclusively 82 female participants in treatment and control conditions. Previous studies have not reached a clear consensus on the existence of gender differences in trust behavior, or, if such differences exist, how they manifest and to what extent. For instance, Croson and Buchan (1999) [18] and others [19-21] found no significant difference in first mover behavior, but they observed that women tended to be more trustworthy than men. A meta-analysis, however, suggests a small effect indicating that men may be slightly more trusting as first movers, while no gender difference was found in trustworthiness [22]. Dittrich (2015) [23] further highlighted a broader limitation in trust game studies: the typical set-up using only university students, which may not be representative of the entire population. In a large-scale experiment with heterogeneous subjects who are representative of the German population, Dittrich identified age-dependent gender differences, noting an inverse U-shape relation between age and trust/reciprocity in men, with no age effects for women. This finding suggests that subject heterogeneity can reveal more nuanced patterns in trust games.

We recognized the exclusively female participant setting as a limitation of this study. Like many trust game experiments—including our own—this study relies on a narrow and homogeneous sample. While this limitation should be considered when interpreting our results, the key behavioral traits the trusts game is designed to measure, such as the trust level and adaptive patterns, are still effectively reflected in the human experiment we used for comparison. For example, results of these female participants revealed profoundly different human behavior in one-shot and repeated games aligned with a mixed population of both genders found in previous studies [24,25], for which we were curious whether this adaptive patterns would be replicated in LLMs. In another study interested in behaviors of LLMs in repeated social interactions, exclusively female samples were also used [2].

All the human participants in this previous experiment [16] were used to compare with the unspecified personas of LLMs. LLMs with selfish and unselfish personas were compared to human participants who were more or less altruistic. We constructed these two subgroups from human participants based on their responses in a dictator game. Each participant was asked to divide 50 monetary units among themselves and a random other participant. Participants who gave less than 5 units to the other participant, we classify as "selfish". This was about 30% of the total group. Participants who gave 15 points or more to the other participant were classified as "unselfish". This was also about 30% of the total group. These two groups are used for comparison with selfish and unselfish personas.

# 3 Results

## 3.1 LLMs in Games: Strategic Features and Resemblance to Humans

We analyzed the responses of LLMs and compared them with human participants in the one-shot and repeated games to examine whether LLMs' responses varied between these two games. We consider both common strategic considerations and to what extent the reactions of LLMs reflected human-like tendencies such as trust and reciprocity.

In trust games, the expression and repayment of trust are operationalized by the sending and returning amounts, respectively. We used proportional amounts for both measures. The sender can send 0, 4, 8, or 12 points out of the initial endowment of 12 points to the receiver, resulting in four proportional values: 0, 0.33, 0.67, and 1. For receivers, we report the proportion of points returned to the sender out of the total amount they could send back after receiving points from the sender. This value could range from 0 to 1. Note that for multi-round games, in Section 2.1 and 2.2, we used the sending and returning proportions of the first round to ensure a fair comparison with the one-shot game. The sending and returning amounts in the other rounds will be covered in the round-by-round dynamics as discussed in Sections 3.3 and 3.4.

First, we examined whether the responses of LLMs were driven by strategic arguments, acting as if they were rational economic agents with the aim to maximize self-interest in a similar way to the traditional views of "homo economicus" with rationality and no other-regarding preferences. In the one-shot game, assuming this type of behavior, the receiver would never return any amount to the sender, as this would decrease their payoff. The sender, anticipating this, would never send any points in the first place. This results in the subgame- perfect Nash equilibrium of zero trust and exchange in the one-shot game, a theoretical benchmark from which real human behavior typically deviates.

A previous study by Xie *et al*. (2024) [15] examined the behavior of large language models (LLMs) when they adopted 53 distinct personas characterized by specific names, genders, addresses and other demographics, and found that the models deviated from rational economic behavior by exhibiting a certain level of trust. Our study shows that LLMs demonstrate trust even without being prompted to assume any specific persona—that is, when operating under an unspecified persona. As shown in Fig. 1, none of the LLMs with unspecified persona responded like purely self-interested rational actors. The exchange amounts were much higher than zero for the three unspecified persona variants as well as the other specific personas of selfish and unselfish players. It suggests that, even in the default or selfish mode of LLMs, their reactions resembled a certain level of trust and reciprocity that is not driven by strategic arguments alone. We compared trust levels by unspecified, selfish and unselfish persona to agent trust in Xie et al. (2024) [15]. Since Bard and Claude were not included in Xie *et al*.

(2024), comparison was made for ChatGPT4. When prompted with 53 personas featuring detailed demographics, ChatGPT4 demonstrated a relatively high level of trust in the one-shot game, with a median sending amount of 70%. This level of trust is comparable to the unselfish persona in our study, but the unspecified or selfish personas of ChatGPT showed much lower trust.

For repeated and the probabilistic repeated games, the rational behavior is more ambiguous because multiple equilibria exist depending on what agents exactly believe about how others will behave. In general, the prediction is that senders and receivers share more points with the other the more rounds of the game they expect to play. This pattern is not systematically reproduced by LLMs. ChatGPT-4 and Claude exchanged more points in the repeated game compared to the one-shot game, but Bard did not align with this theoretical expectation.

We also compared the trust level reflected in LLMs' responses to that of humans. In the one-shot game, we found no differences in the proportions sent by humans and the default modes of three LLMs (Fig. 1a). It suggests LLMs managed to emulate how human players established trust at the very beginning of a one-shot game. Trust in the game is initialized when an agent performs an initial sacrifice—that is, a certain amount given away by the sender without any control over the return. Humans and LLMs all gave away approximately half of their initial points to the receiver as the first gesture of trust. From the same starting point, however, LLMs and humans diverged as to how they repaid the trust. LLMs paid back significantly higher amounts than humans, with a returning amount between 23% to 32% for LLMs compared to 17% for humans.

In the repeated game, only Claude, in its default mode, emulated how human players established trust at the beginning of the game by sending over 60% to the receiver (Fig. 1b), a percentage that was higher than what they did in the one-shot game (~50%). The percentages sent by ChatGPT-4 and Bard, in contrast, were around 50%, close to what they sent in the one-shot game. How LLMs repaid trust in the repeated game significantly differed from humans. The returning amount by ChatGPT-4 (51%) was higher than that of humans (34%), whereas Bard and Claude returned significantly lower amounts between 17% and 21%. As such it appears that more human-machine differences emerge in the repeated game where decision-making becomes more complex and long-term trust can be developed.

It should be noted that, in the one-shot game, where we did not find significant human-machine differences in how they initialized the game, some small but consistent differences were observed across LLMs. As the sender, the amount sent by Claude was 8-12% more than those from ChatGPT-4 and Bard. The non-significant differences between humans and machines, contrasted with significant differences across models, may be attributed to varying sample sizes and greater variability in human responses. The human data sample size was 492, smaller than that of each LLM (n = 630~700). Additionally, the standard deviation in human responses was twice that of each LLM. This implies that despite the smaller sample size and the clustering of observations within 82 participants, there was more variance in human responses. For each LLM, the standard deviation reflected the variability in its own responses. The above results suggest an interesting dynamic that while all LLMs provided human-like responses, they were uniquely different, which might be due to the differences in their training datasets or underlying neural networks.

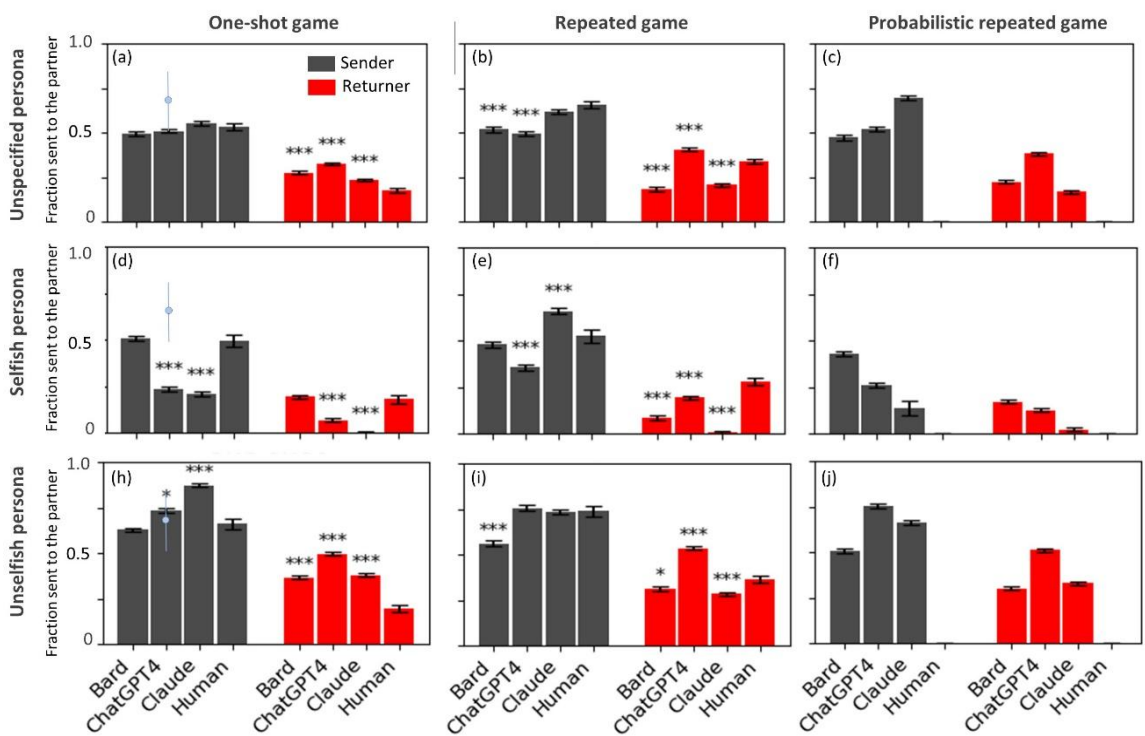

**Figure 1**: Fraction of points exchanged by LLMs and human participants. The vertical axis represents the averaged proportion of points sent (black bars) or returned (red bars), with error bars indicating standard errors. The responses of three large language models (LLMs)—-Bard, ChatGPT-4, and Claude—-are compared across three personas: unspecified (default), selfish, and unselfish. The blue dot and line above ChatGPT-4 represent results from Xie et al. (2024) [15] using the same LLM but featuring 53 personas with detailed demographics, with the dot indicating the median amount sent and the line spanning the 25th to 75th percentiles. An asterisk indicates a significant difference between the factions by LLMs and humans. The significance levels are marked as follows: *** for p-values less than 0.001, ** for p-values between 0.001 and 0.01, and * for p-values between 0.01 and 0.05.

## 3.2 LLMs: Selfishness and Unselfishness

As the unspecified persona of LLMs did not always follow strategic arguments for behavior neither always resembled human behavior, we further examined how LLMs adapt to contrasting selfish and unselfish personas and the trust level reflected in these adapted responses.

With a few exceptions of Bard and ChatGPT-4, the adoption of selfish and unselfish personas generally triggered distinct responses from those of unspecified personas. Using responses of unspecified persona as baselines, LLMs generally lowered the amount exchanged as selfish players and increased the amount as unselfish players. This generally aligned with the expected patterns for each persona and the expectation that the unspecified persona revealed the natural or moderate approach taken by LLMs.

LLMs, however, diverged on the exact amounts exchanged as selfish players. As selfish senders, the amounts sent by Bard were similar to those of humans, which was approximately 49% and 52% in the one-shot and repeated games, respectively. Notably, those amounts were not only positive but significantly greater than zero. Suggesting that, even for selfish players, their response deviated significantly from strategic self-interested actors. For ChatGPT-4 and Claude, the sending amounts were also above zero. A selfish ChatGPT-4 player sent out 23% and 35% in the one-shot and repeated games, respectively. Both were lower than the amounts by Bard and humans. The response of Claude as a selfish sender was not consistent. It transferred 20% in the one-shot game, but the amount in the repeated game was as high as 66%. In the probabilistic repeated game, which was used to check the consistency of LLMs' responses in multi-round games, the amount sent by Claude dropped dramatically to 18%.

As selfish receivers, all LLMs unanimously returned much lower amounts than humans did. In contrast to the 18% returned by humans in the one-shot game, Claude, ChatGPT-4, and Bard paid back 0%, 7%, and 17% respectively. In the repeated game, humans returned 28%, whereas Claude, Bard, and ChatGPT-4 sent back 0%, 9%, and 19% respectively. With a returning amount close to zero, a selfish returner played by Claude responded in a way similar to a strategic self-interested actor.

In regards to unselfish senders, Bard gave away 66% in the one-shot game, similar to the choice made by unselfish human players. The amounts sent by unselfish ChatGPT-4 and Claude were much higher, which was 73% and 87%, respectively. In the repeated game, the amounts sent by unselfish players of humans, ChatGPT-4 and Claude, were similar, around 75%, whereas Bard sent 56%. Acting as unselfish receivers, the amounts paid back by three LLMs (37-50%) were significantly more than humans (20%) in the one-shot game. An unselfish player of ChatGPT-4 returned 53% in the repeated game, more than the amounts paid back by humans (37%), Bard (32%) and Claude (29%).

Compared to the unspecified persona, the differences in the exact amounts exchanged between LLMs and humans, and among LLM systems were slightly larger when particular persona such as selfish and unselfish were specified. Some human-machine alignments were found in the expression of trust as selfish or unselfish senders, but there were minimal similarities in how humans and LLMs when positioned as receivers, reciprocated trust under selfish or unselfish personas.

## 3.3 Interactive Reasoning of LLMs

Decisions in the trust game are influenced by social preferences and players' reasoning about the game's structure and their counterparts' actions. Here we examined how LLMs responded to changes in game mechanics and the actions of their counterparts—a capability we refer to as interactive reasoning. In contrast to one-shot, players in repeated games can adapt their behavior based on their partner's behavior before. An adaptive player would become more or less trusting or trustworthy depending on how trustworthy or trusting their partner was before. For example, a sender would increase the amount sent in the next round if a receiver repays generously. An interactive reasoning pattern that can be termed as adaptive behavior.

Human participants demonstrate adaptative behavior and reciprocal responses to their counterparts' actions. Compared to one-shot game, they adopt distinct strategies in the repeated game by sending and returning significantly higher amounts (Fig.2d). In addition, human participants rewarded or punished the generous or self-interested behaviors of their counterparts. Reciprocity expressed by one player strongly predicts future trust expressed by their partner, and vice versa [24-26]. Analyzing the responses from all human participants in the experiments, if the proportion returned by the receiver increases by 10%, the sender in the next round would increase the amount sent by about 14%. Conversely, if the sender gives 10% more, the receiver would also increase the payback amount by about 3%.

We examined whether the responses of LLMs in their default positions showed traits of interactive reasoning as evident in humans. First, only Claude strategically distinguished its responses in the one-shot and repeated games by elevating the sending and returning amounts in the latter. Responses of ChatGPT-4 and Bard were largely insensitive to the changes in game structures, with one exception of ChatGPT-4 which increased the returning amount in the repeated game. Second, regarding the reciprocal adoption of the actions of counterparts, this pattern was mostly absent in the repeated games played by LLMs, with the exceptions of Claude and Bard as the receivers. Claude and Bard slightly adjusted the proportion they returned in the repeated game in response to the higher trust expressed by the sender. If the proportion given by the sender increased by 10%, the percentage by the receiver would increase by 1% and 0.3% for Claude and Bard, respectively. This reciprocal adaption effect, however, was much smaller compared to that of human participants.

To validate the consistency in LLMs' responses in multi-round games, a close variant of the repeated game was introduced– the probabilistic repeated game. It has a similar set-up as the repeated game, but a six-faced dice is thrown at the end of each round, which ends the game prematurely if a six is rolled. As such, uncertainty is introduced, and the complexity of decision-making is increased in the probabilistic repeated game, though we still expect the pattern of multi-round interactive reasoning to be retained in this game, namely in the form of higher sending and returning amounts compared to the one-shot game and reciprocal adaptation to the actions of the other players. As shown in Table 2, the absence of adaptive behavior by ChatGPT-4 and Bard in the repeated game persisted in the probabilistic repeated game. And the adaptive behavior found in Claude in the repeated game disappeared. Reciprocal adaption was also not consistent with the findings in the repeated game. The inconsistencies suggest that the occasional observed interactive reasoning in LLMs' unspecified responses is likely random rather than a stable, reproducible characteristic.

Another notable pattern that emerged from human responses is that selfish and unselfish participants retain similar traits of interactive reasoning in their responses. Adaptive behavior dependent on experiences with partners occurs across selfish and unselfish participants. In addition, the magnitudes of reciprocal adaption were similar for both selfish and unselfish human players, as shown in Table 1. When acting as senders, selfish and unselfish human players increased the amount they sent by 13.8% and 13.1% in response to every 10% increase in the reciprocity shown by the other players, respectively. As receivers, these effects were 3.3% and 3.2% for selfish and unselfish players, respectively. Different human personas were not associated with distinct reciprocal adaptation patterns.

As adaption behavior was missing in unspecified Bard players, this pattern persisted in the selfish and unselfish personas of Bard. Different personas in Claude and ChatGPT-4, on the other hand, triggered different patterns of adaptive behavior. For ChatGPT-4, more alignment with human adaptive behavior was found in both selfish and unselfish personas, and these adaption patterns were stable in both multi-round games. The opposite was found for Claude, in which adaptive behavior became absent when it acted as selfish and unselfish players. Regarding reciprocal adaption, only ChatGPT-4 showed clearer patterns of reciprocal adaptation when given personas, especially in the probabilistic repeated games. As shown in Table 1, positive and statistically significant correlations were found between the amounts sent and returned for both the selfish and unselfish personas of ChatGPT-4. The positive effect observed in ChatGPT-4's responses regarding the returning amount by unselfish persona in relation to the sending amount was comparable to that of humans, making it the closest resemblance of human reciprocal adaption observed in this study. In short, Bard and Claude's responses under different personas did not show increased stability or consistency in their interactive reasoning. In contrast, ChatGPT-4's patterns of interactive reasoning appeared more stable and consistent when personas were given.

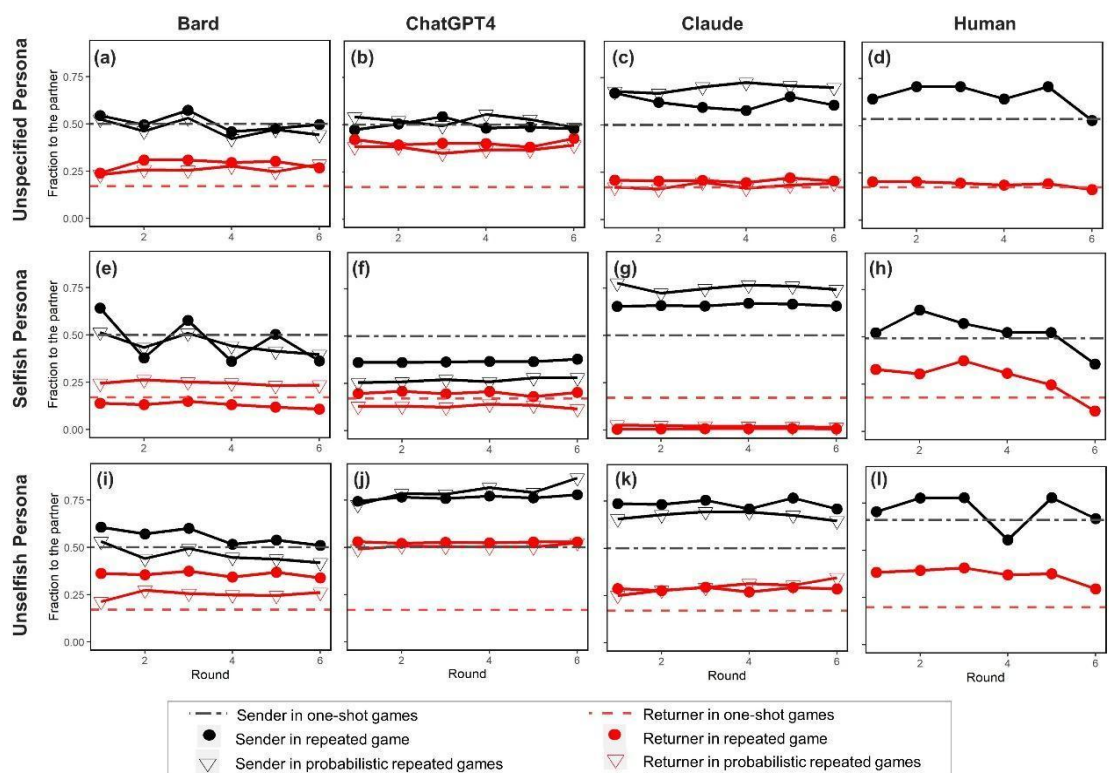

**Figure 2**: Round-by-round responses by LLMs and human participants in repeated games and probabilistic repeated games, with responses of one-shot games showing as references (dotted lines).

**Table 1** Correlation coefficients between the proportions sent and returned by LLMs and humans [a]

| Game | Agents | The sending amount ~ the returning amount | | | The returning amount ~ the sending amount | | |
|---|---|---|---|---|---|---|---|
| | | Unspecified | Selfish | Unselfish | Unspecified | Selfish | Unselfish |
| Repeated game | Human | 1.43 *** | 1.38 *** | 1.31 *** | 0.32 *** | 0.33 *** | 0.32 *** |
| | Bard | -0.05 | -0.21 *** | 0.21 *** | 0.03 * | 0.004 | 0.14 *** |
| | ChatGPT4 | 0.03 | 0.113 ** | 0.04 | 0.02 | -0.05 *** | 0.03 |
| | Claude | -0.23 *** | -0.22 | -0.26 *** | 0.1 *** | -0.003 | 0.04 *** |
| | Bard | 0.07 * | 0.13 *** | 0.21 *** | -0.15 *** | -0.14 *** | -0.04 * |
| | ChatGPT4 | 0.003 | 0.13 ** | 0.157 *** | 0.11 *** | 0.16 *** | 0.28 *** |
| Probabilistic repeated game | Claude | -0.14 *** | -0.1 | -0.24 *** | 0.06 *** | 0.007 | -0.13 *** |

[a] The significance levels are marked as: *** for p-values less than 0.001, ** for p-values between 0.001 and 0.01, and * for p-values between 0.01 and 0.05.

## 3.4 Interaction of Persona, Game Structure, and LLMs in Shaping Responses

We observed varied game outcomes in terms of sending and returning responses in the previous discussion. Here, we aim to determine whether these differences arise primarily from the type of LLMs, the type of personas, the type of games, or their interactions.

None of the above factors alone can explain the variability of game responses, but the type of personas appears to have the most substantial impacts, followed by the type of LLMs and the type of games (See Appendix Marginal Effect Analysis). The differences in game responses (the proportions sent and returned) relevant to the change of persona were approximately 2 to 2.5 times greater than those resulting from the type of LLMs. This suggests role-playing or persona-setting significantly shapes the responses of LLMs. This framing effect is much more important than the specific model being used. Regarding the effect associated with changes in games, its effect was nearly negligible in explaining the returning responses. A small effect from the type of games was found in the sending proportion, which was equal to 18% of the effect relevant to the change of personas.

Strong and non-linear interaction effects between these three factors were found, primarily in the sending responses (See Appendix Interaction Model). Using the unspecified persona, ChatGPT-4, and repeated games as reference levels, we found different patterns for the sending and returning responses. For returning responses, the main effect of LLM type (changing from ChatGPT-4 to Bard or Claude) and persona type (changing from unspecified to selfish or unselfish) explained most of the variations, with only weak interaction effects between LLMs, personas, and games. However, for sending responses, while the main effect of persona remained strong, the interaction effects between LLMs, persona, and games were equally substantial, reaching a comparable level to the main effect of persona. It suggests while persona remained a major factor, the variations in sending responses were also substantially influenced by contextual factors, specifically how different LLMs responded to particular game settings.

## 4 Conclusion and Discussion

While a primary goal of LLMs is to provide human-like responses in conversations—a goal they largely achieve—we focused our investigation on how LLMs perform in behavioral contexts where they cannot easily derive straightforward answers from their training data. By placing LLMs in different contexts of trust and reciprocity, we investigated the emerged responses and studied how similar they were to human behaviors in comparable contexts. Results from this study provide proxies to fathom the extent to which LLMs can provide human-like behavioral responses in social exchange. However, caution should be taken as responses here reflect patterns in data, and it remains unclear how close they were to autonomy and conscious decision-making.

We decoded social interactions into two facets: social preferences and interactive reasoning. Social preferences refer to how individuals value their own payoffs in relation to the outcomes of others, relevant to the inherent sense of fairness, trust, and reciprocity. Interactive reasoning captures how participants react and learn from changes in interaction contexts, such as game settings and actions of others, to reinforce their previous strategies or adapt to new ones. These two facets of responses of Bard, ChatGPT-4, and Claude were analyzed and summarized in Table 2.

**Table 2** Summary of the LLMs responses

| LLMs | Personas | Players | Interactive reasoning | | | Resembling human trust level [c] | |
|---|---|---|---|---|---|---|---|
| | | | One-shot < Repeated [a] | One-shot < Probabilistic Repeated [a] | Reciprocal adaptation [b] | One-shot | Repeated |
| Bard | Unspecified | Sender | | | ◐ | ● | -21% |
| | | Returner | | | ◑ | 57% | -47% |
| | Selfish | Sender | | | ◐ | ● | ● |
| | | Returner | | | | ● | -68% |
| | Unselfish | Sender | | | ● | ● | -24% |
| | | Returner | | | ◑ | 85% | -14% |
| ChatGPT4 | Unspecified | Sender | | | | ● | -24% |
| | | Returner | ● | ● | ◐ | 86% | 50% |
| | Selfish | Sender | ● | ● | ● | -53% | -32% |
| | | Returner | ● | ● | ◐ | -61% | -32% |
| | Unselfish | Sender | ● | ● | ◐ | 11% | ● |
| | | Returner | ● | ● | ◐ | 150% | 47% |
| Claude | Unspecified | Sender | ● | | | ● | ● |
| | | Returner | ● | | ● | 34% | -38% |
| | Selfish | Sender | ● | | | -59% | 27% |
| | | Returner | | | | -101% | -97% |
| | Unselfish | Sender | | | | 32% | ● |
| | | Returner | ● | ● | ◑ | 90% | -22% |

[a] ● indicates such a pattern is found.
[b] Reciprocally adapt to counterparts' actions (i.e., positive correlations between the sending and returning proportions) in the repeated game (◑), probabilistic repeated game (◐) or both games (●).

All three LLMs made choices in line with social preferences that distinguished them from idealized self-interested actors with perfect rationality, when even no specific persona was prompted. In the most simple game setting characterized by a one-shot interaction, Bard, ChatGPT-4, and Claude sent amounts to the other player comparable to human trust behavior [22]. Responses of the three LLMs were all human-like, but there were small yet consistent differences across models. If we assume the training data of these LLMs were largely similar, this suggests different architectures in neural networks (such as layers, nodes, and weights) can still produce unique representations of data patterns in social interactions. We also need to acknowledge that the resemblance was largely on with average human behavior, while the distributions of behavior showed considerably more variance among humans than with the different LLMs.

As the games involved the repayment of trust or multi-round interactions with the same player, responses provided by LLMs were rarely human-like. Little alignment between LLMs and humans was found in the repayment of trust in the one-shot interaction, even though they started with similar expressions of trust. Facing the same offer by the sender in the one-off interaction, ChatGPT-4, Bard, and Claude paid back more than humans did. However, this does not imply that LLMs were always more trustworthy than humans, as the relative differences between humans and LLMs in trust repayment varied across different games and personas. If they have interacted with the same player for multiple rounds, responses of LLMs differed from humans in both how they established and repaid trust. The initial sacrifices made by LLMs were generally lower than those offered by humans (with the exception of Claude), and the same pattern was observed in the repayment of trust (with the exception of ChatGPT-4).

On the one hand, these human-machine divergences indicate LLMs and humans had different social preferences regarding how to repay trust. In one-shot interactions, LLMs responded as if they were more generous than humans in the amount they paid back. However, this pattern was context-specific and did not consistently appear in other situations, such as in multi-round games. On the other hand, the differences between humans and LLMs emerged as human participants changed their strategies from one-shot interaction to multi-round games, whereas LLMs like Bard and ChatGPT-4 did not sensibly adjust their responses to the changes of game settings. As human participants significantly increased the amounts they sent and returned in multi-round games compared to one-shot interactions, LLMs such as Bard and ChatGPT-4 either exchanged similar amounts or even lower amounts in the multi-round games, indicating an absence of including strategic arguments in their responses. This resulted in widening gaps between the responses of humans and LLMs.

While human participants adapt their behavior strongly dependent on what their partners do, such reciprocal adaptive behavior was not consistently observed in LLMs' responses. There were not many variations in the amounts sent and returned by LLMs in different rounds compared to the variability seen in human participants. LLMs such as Claude and Bard occasionally adopt the amount they sent or returned in reaction to the amount given by their counterparts, but this adaption effect was much smaller than that of humans and was not stable and reproducible across contexts. While existing studies have emphasized the evolutionary breakthrough of LLMs like ChatGPT in cognitive and reasoning tasks [27–30], these models might excel only in static and well-defined scenarios. The limited extent of adaptive behavior suggests current LLMs still have difficulty reproducing effective interactive reasoning to navigate the complex dynamics underlying real-world social interactions, where participants actively sense the changes in the environment and the actions of other players and factor these into their decisions.

We found a substantial framing effect on the responses of LLMs when they were prompted to behave as selfish or unselfish persona. The type of persona accounted for the most variation in LLMs' responses, having a larger impact than that caused by the choice of different LLMs (Bard, ChatGPT-4, and Claude) or the specific games played. By giving a specific persona, ChatGPT-4 showed more stable reproduction of strategic adaption and reciprocal adaption. While a previous study suggested LLMs were sensitive to the game structures and surrounding contexts [14], this study highlighted the even more profound effect caused by the framing of persona. In addition, this significant framing effect was driven by the portrayal of personas as selfish or unselfish, rather than by depiction of detailed demographic information [15]. It opens up a discussion on what types of information should be specified to define LLM personas for different purposes—such as generating representative samples for surveys, behavioral experiments, observational studies, and more.

## Declaration of generative AI and AI-assisted technologies in the writing process

During the preparation of this work, the authors used ChatGPT, Claude, and Grammarly to improve language. After using these tools, the authors reviewed and edited the content as needed and thus take full responsibility for the content of the publication.

## References

1. Grossmann, I. et al. AI and the transformation of social science research. *Science,* 380, 1108–1109 (2023).
2. Akata, E., Schulz, L., Coda-Forno, J. et al. Playing repeated games with large language models. *Nature Human Behaviour,* https://doi.org/10.1038/s41562-025-02172-y (2025).
3. Liu, Z., Anand, A., Zhou, P., Huang, J., Zhao, J. InterIntent: Investigating Social Intelligence of LLMs via Intention Understanding in an Interactive Game Context. https://doi.org/10.48550/arXiv.2406.12203 (2024).
4. Turing, A. M. Computing Machinery and Intelligence. *Mind, New Series,* , 433–460 (1950).

5. Binz, M. & Schulz, E. Using cognitive psychology to understand GPT-3. *Proceedings of the National Academy of Science*s 120, e2218523120 (2023).

6. Lampinen, A. K. et al. Language models, like humans, show content effects on reasoning tasks. *PNAS Nexus* 3, 233. 2752-6542 (2024).

7. Hagendorff, T., Fabi, S. & Kosinski, M. Human-like intuitive behavior and reasoning biases emerged in large language models but disappeared in ChatGPT. *Nature Computational Science* 3, 833–838 (2023).

8. Douglass, R. W., Gartzke, E., Lindsay, J. R., Gannon, J. A. & Scherer, T. L. What is Escalation? Measuring Crisis Dynamics in International Relations with Human and LLM Generated Event Data. https://arxiv.org/abs/2402.03340 (2024).

9. Lamparth, M. et al. Human vs. machine: Language models and wargames. https://doi.org/10.48550/ arXiv.2403.03407 (2024).

10. Rivera, J. et al. Escalation risks from language models in military and diplomatic decision-making. https://doi.org/10.1145/3630106.3658942 (2024).

11. Aher, G. V., Arriaga, R. I. & Kalai, A. T. Using Large Language Models to Simulate Multiple Humans and Replicate Human Subject Studies. https://arxiv.org/abs/2208.10264 (2023).

12. Nowak, M. A. & Sigmund, K. Evolution of indirect reciprocity. *Nature* 437, 1291–1298 (2005).

13. Johnson, T. & Obradovich, N. Measuring an artificial intelligence language model's trust in humans using machine incentives. *Journal of Physics: Complexity 5*, 015003 (2024).

14. Lorè, N. & Heydari, B. Strategic behavior of large language models and the role of game structure versus contextual framing. *Scientific Reports* 14, 18490 (2024).

15. Xie, C., Chen, C., Jia, F., et al. Can Large Language Model Agents Simulate Human Trust Behavior? 38th Conference on Neural Information Processing Systems (NeurIPS 2024).

16. Buskens, V., Raub, W., Miltenburg, N., Montoya, E. & van Honk, J. Testosterone Administration Moderates Effect of Social Environment on Trust in Women Depending on Second-to-Fourth Digit Ratio. *Scientific Reports*, 6, 27655 (2016).

17. Wei, J., Wang, X., Schuurmans, D., et al. Chain of thought prompting elicits reasoning in large language models. NIPS'22: Proceedings of the 36th International Conference on Neural Information Processing Systems, 1800: 24824 - 24837 (2022).

18. Croson, R., Buchan, N. Gender and Culture: International Experimental Evidence from Trust Games. *American Economic Review* 89 (2): 386–391 (1999).

19. Clark, K., Sefton, M. The Sequential Prisoner's Dilemma: Evidence on Reciprocation. *Economic Journal*, 111, 51-68 (2001).

20. Cox, J., C., Deck, C., A. When are Women More Generous than Men? *Economic Inquiry*, 44, 587-598 (2006).

21. Chaudhuri, A., Gangadharan, L. An Experimental Analysis of Trust and Trustworthiness. *Southern Economic Journal*, 73, 959-985 (2007).

22. van den Akker, O., van Assen, M., van Vugt, M., Wicherts, J. Sex differences in trust and trustworthiness: A meta-analysis of the trust game and the gift-exchange game. *Journal of Economic Psychology*, 81, 102329 (2020).

23. Dittrich, M. Gender differences in trust and reciprocity: evidence from a large-scale experiment with heterogeneous subjects. *Applied Economics*, 47(36), 3825–3838 (2015).
24. Camerer, C., Weigelt, K. Experimental Tests of a Sequential Equilibrium Reputation Model. *Econometrica*, 56 (1): 1-36 (1988).
25. Neral, J., Ochs, J. The Sequential Equilibrium Theory of Reputation Building: A Further Test. *Econometrica*, 60 (5): 1151-1169 (1992).
26. King-Casas, B., Tomlin, D., Anen, C, et al. Getting to know you: Reputation and trust in a two-person economic exchange. *Science* 308, 78–83 (2005).
27. Brown, T., Maan B., Ryder, N., et al. Language models are few-shot learners. NIPS'20: Proceedings of the 34th International Conference on Neural Information Processing Systems, 159, 1877 – 1901 (2020).
28. Bubeck, S., Chandrasekaran V., Eldan, R., et al. Sparks of artificial general intelligence: Early experiments with GPT-4. https://doi.org/10.48550/arXiv.2303.12712 (2023).
29. Kojima, T., Gu, S. S., Reid, M., Matsuo, Y. & Iwasawa, Y. Large language models are zero-shot reasoners. NIPS'22: Proceedings of the 36th International Conference on Neural Information Processing Systems: 1613, Pages 22199 - 222 (2024)
30. Azaria, A., Mitchell, T. The Internal State of an LLM Knows When It's Lying. https://doi.org/10.48550/arXiv.2304.13734 (2023).